\documentclass[letterpaper]{article} 
\usepackage{aaai2026}  
\usepackage{times}  
\usepackage{helvet}  
\usepackage{courier}  
\usepackage[hyphens]{url}  
\usepackage{graphicx} 
\usepackage{natbib}  
\usepackage{caption} 
\usepackage{algorithm}
\usepackage{algorithmic}

\usepackage{newfloat}
\usepackage{listings}
\DeclareCaptionStyle{ruled}{labelfont=normalfont,labelsep=colon,strut=off} 
\floatstyle{ruled}
\newfloat{listing}{tb}{lst}{}
\floatname{listing}{Listing}

\usepackage{algorithm}
\usepackage{algorithmic}
\usepackage{amsmath}
\usepackage{amssymb}
\usepackage{booktabs}
\usepackage{multirow}
\usepackage{bm}

\usepackage{xcolor}

\title{Learning Diffusion Policy from Primitive Skills for Robot Manipulation}
\author{
    Zhihao Gu\textsuperscript{\rm 1}\thanks{Work done when Zhihao Gu was a Postdoc of Prof. Dong Xu.},
    Ming Yang\textsuperscript{\rm 2},
    Difan Zou\textsuperscript{\rm 1},
    Dong Xu\textsuperscript{\rm 1}\equalcontrib
}
\affiliations{
    \textsuperscript{\rm 1}Department of Computer Science, School of Computing and Data Science, The University of Hong Kong\\
    \textsuperscript{\rm 2}School of Software, Beihang University\\
    ellerygzh@gmail.com, viv@buaa.edu.cn, \{dzou, dongxu\}@cs.hku.hk

}

\usepackage{bibentry}

\begin{document}

\maketitle

\begin{abstract}
Diffusion policies (DP) have recently shown great promise for generating actions in robotic manipulation. However, existing approaches often rely on~\textit{global} instructions to produce~\textit{short-term} control signals, which can result in misalignment in action generation. We conjecture that the primitive skills, referred to as fine-grained, short-horizon manipulations, such as ``move up'' and ``open the gripper'', provide a more intuitive and effective interface for robot learning. To bridge this gap, we propose SDP, a skill-conditioned DP that integrates interpretable skill learning with conditional action planning. SDP abstracts eight reusable primitive skills across tasks and employs a vision-language model to extract discrete representations from visual observations and language instructions. Based on them, a lightweight router network is designed to assign a desired primitive skill for each state, which helps construct a single-skill policy to generate skill-aligned actions. By decomposing complex tasks into a sequence of primitive skills and selecting a single-skill policy, SDP ensures skill-consistent behavior across diverse tasks.
Extensive experiments on two challenging simulation benchmarks and real-world robot deployments demonstrate that SDP consistently outperforms SOTA methods, providing a new paradigm for skill-based robot learning with diffusion policies.
\end{abstract}


\section{Introduction}
Enabling robots to perform diverse real-world tasks has been a long-standing goal in robotics and artificial intelligence.
One promising approach is to teach robots by example, allowing them to learn directly from demonstrations. However, it is uniquely challenging: unlike standard prediction problems, robotic control demands precise, context-aware actions~\cite{chi2023diffusion}.
To handle it, prior research focused on improved action representations~\cite{mandlekar2021matters,shafiullah2022behavior} and richer internal models of robot behavior~\cite{florence2022implicit,wu2020spatial}.

\begin{figure}
     \centering
    \includegraphics[width=0.99\linewidth]{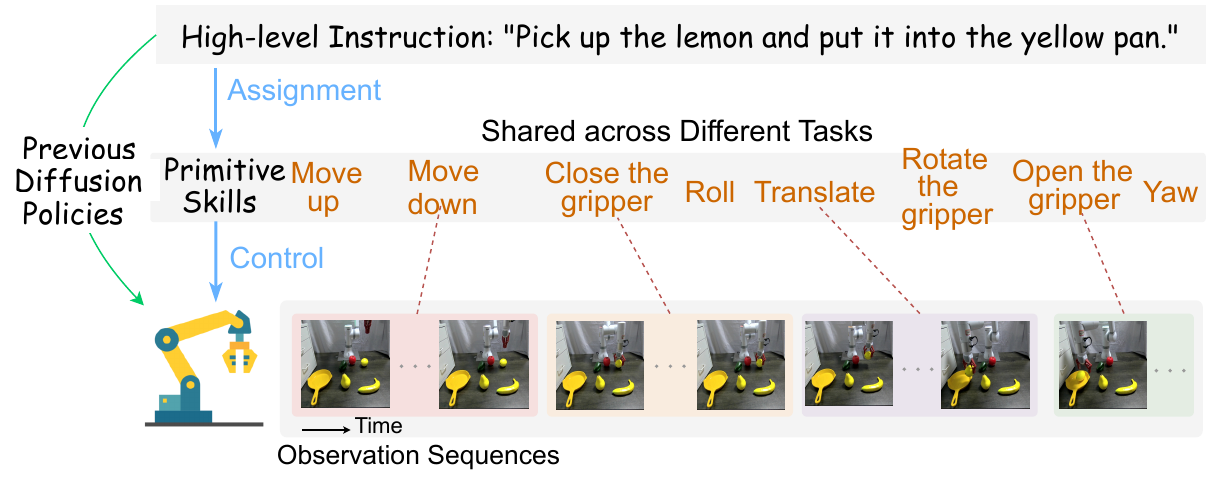} 
    \caption{A task consists of a series of short-term manipulations, and we abstract them into eight shared primitive skills, which provide concrete instruction. Previous diffusion policies map high-level instructions to actions directly. In contrast, we learn primitive skills and integrate them into the conditional action generation for more precise control.}\label{motiva}
\end{figure}

Recently, diffusion models~\cite{ho2020denoising,mokady2023null}, a class of generative models that learn to reverse a gradual noise-adding process, have achieved remarkable success in high-fidelity image generation~\cite{ho2020denoising,jo2022score,rombach2022high}. Building on this success, diffusion models have been explored in robotics for generating action sequences. The Diffusion Policy (DP)~\cite{chi2023diffusion} is a pioneering work that generates robot behavior via the conditional denoising process of diffusion models.
Instead of directly outputting an action, it infers the action-score gradient for several denoising iterations, significantly improving the performance. However, conditioned only on visual observations, it is difficult to learn multiple tasks at the same time, severely limiting its deployment in real-world scenarios. 

To overcome these limitations, recent work has introduced natural language instructions as a condition, enabling robots to perform a broader range of tasks~\cite{ha2023scaling,reuss2407multimodal,liu2024rdt,wang2407sparse,reuss2024efficient}. Typically, a language encoder transforms instructions into embeddings, which, together with noisy action sequences, are sent to the diffusion policy for action generation. Research in this paradigm has advanced along three dimensions~\cite{song2025survey}: robot data representations~\cite{wang2407sparse,ze20243d,wang2024scaling}, model architectures~\cite{mees2024octo,reuss2407multimodal,ye2024latent}, and diffusion strategies~\cite{ren2024diffusion,reuss2407multimodal,liang2024skilldiffuser}. Despite these advances, most existing methods map high-level instructions directly to short-term actions, which can result in ambiguous or misaligned behaviors. For example, if the robot is going to close the gripper, the high-level task description, ``Pick up the lemon and put it into the pan'', would be too abstract to provide an explicit instruction, while a more fine-grained level of instruction should be included. This motivates us to design a method that can~\emph{generate concrete short-term instructions (named primitive skills)}, such as ``close the gripper'', and~\emph{learn single-skill diffusion policies} to generate more accurate actions.

\begin{figure}
     \centering
    \includegraphics[width=0.95\linewidth]{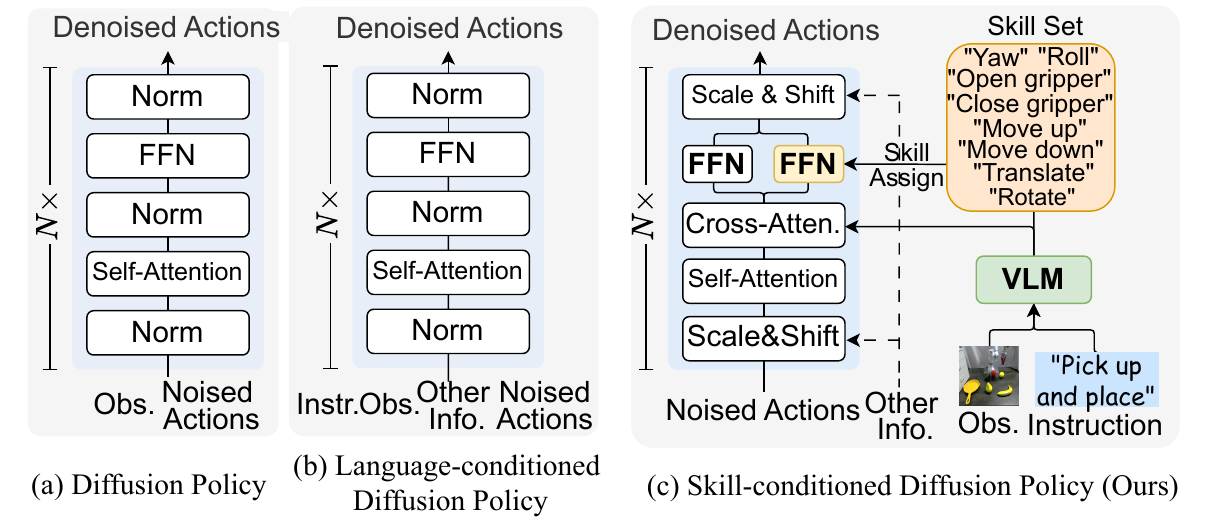} 
    \caption{Comparison between (a) diffusion policy (DP), (b) language-conditioned DP, and (c) our skill-conditioned DP. Ours executes abstract instructions with more precise guidance from the assigned primitive skills.}\label{compari}
\end{figure}

To address this gap, we propose SDP, a skill-conditioned diffusion policy that combines fine-grained skill learning with conditional, low-level action generation. Our approach is built on two key ideas: (1) decomposing ambiguous, high-level instructions into learnable short-term skills based on current observations, and (2) training a diffusion policy that generates actions conditioned on these skills. Specifically, we first abstract short-term manipulations across various tasks into eight primitive skills (see Figure~\ref{motiva}), which in turn can be composed to form complex tasks, and convert visual observations and high-level instructions into discrete representations by a vision-language model. A lightweight router network then dynamically assigns the appropriate skill for each state. The assigned skill synthesizes parameters of the feed-forward network (FFN) in diffusion policy, while additional information, such as proprioception, is injected using an AdaLN operation. The resulting single-skill diffusion policy is capable of producing coherent and precise behaviors aligned with the skill. Compared to previous approaches (see Figure~\ref{compari} (a) and (b)), our SDP interprets and executes complex instructions end-to-end with greater accuracy and more precise control. Extensive experiments on two challenging simulation benchmarks and real-world robot deployments demonstrate the superior performance of SDP. In summary, our main contributions are as follows:
\begin{itemize}
    \item We present SDP, a skill-conditioned diffusion policy that combines fine-grained skill learning and low-level action generation, to mitigate the misalignment in granularity between global instruction and short-term actions.
    \item We introduce eight reusable primitive skills that generalize across diverse manipulation tasks, providing a structured and interpretable action space for robot learning. Furthermore, to leverage these skills effectively, we design a lightweight router network that dynamically assesses state relevance and selects the optimal skill, ensuring adaptive and task-aligned behavior generation.
    \item We design a novel single-skill diffusion policy that generates actions precisely aligned with each skill. By dynamically parameterizing the policy's FFN layer from the assigned skill, we can effectively capture the dependency between primitive skills and low-level control signals.
    \item We demonstrate better multi-task and generalization capabilities than baselines across simulated and real-world tasks. The visualization of skills also reveals its ability to decompose abstract instructions and compose primitive skills, validating its effectiveness and interpretability.
\end{itemize}

\section{Related Work}

\subsubsection{Diffusion policy in robot Manipulation.} 
Diffusion models~\cite{ho2020denoising,mokady2023null} have recently achieved remarkable success in a variety of fields, and their potential for robotic manipulation has attracted growing interest. In the robotics community, researchers have focused on developing diffusion-based policies that enable robots to follow language instructions and perform diverse tasks~\cite{ha2023scaling,reuss2407multimodal,liu2024rdt,wang2407sparse,reuss2024efficient}. These efforts span three main areas: robot data representations, model architectures, and diffusion strategies. The data representations include 2D trajectories~\cite{wang2407sparse}, 3D point clouds~\cite{ze20243d}, and combinations of sensory inputs~\cite{wang2024scaling}. Model architectures often combine diffusion models with large language models~\cite{mees2024octo}, transformers~\cite{reuss2407multimodal}, or variational autoencoders~\cite{ye2024latent}. Additionally, different training strategies have been explored, such as integrating reinforcement learning~\cite{ren2024diffusion}, self-supervised learning~\cite{reuss2407multimodal}, and classifier guidance~\cite{liang2024skilldiffuser,mete2024quest}. In contrast, our SDP emphasizes learning executable skills and training a skill-conditioned diffusion policy for more precise and coherent robot control.

\begin{figure*}[t]
\centering
\includegraphics[width=0.8\textwidth]{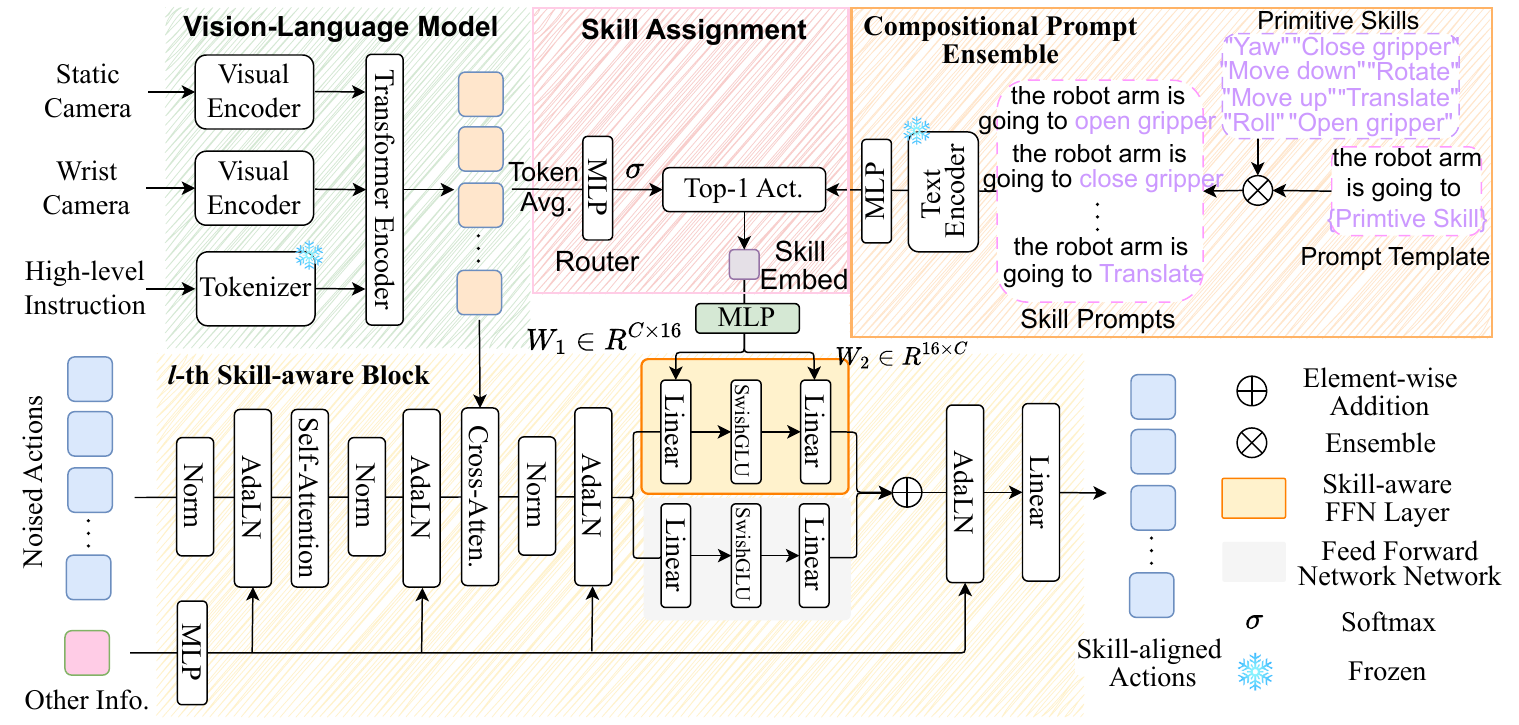}
\caption{Overview of the proposed skill-conditioned diffusion policy (SDP). SDP abstracts short-term manipulations across different tasks into eight primitive skills and introduces a unified prompt template to specify the upcoming manipulations. A router is then designed to assign importance scores for all candidate skills based on the embedding $\bm{z}_{vl}$, generated from visual observations and language instructions by a VLM. Furthermore, a skill with the highest score is selected, which parameterizes an additional FFN layer in the diffusion policy. Other information, such as proprioception, is encoded by an MLP and further injected via an AdaLN operation, resulting in a single-skill policy that predicts skill-aligned actions for precise control.}
\label{framework}
\end{figure*}

\subsubsection{Planning by VLM.} 
Decomposing complex instructions into manageable sub-goals helps robots complete sophisticated tasks more reliably~\cite{dalal2021accelerating,dhakan2022concurrent,hiranaka2023primitive,liu2025skill}. However, manually annotating these sub-goals is labor-intensive and does not scale well. To overcome this, recent methods~\cite{singh2022progprompt,zhang2023bootstrap,ni2024generate} leverage VLMs rich in real-world knowledge to automatically generate task plans for robot learning. Other methods like~\cite{garg2022lisa}, learn codebooks of sub-goals from latent variables, where each code may correspond to multiple states. Our SDP differs in two key aspects. First, we propose a set of human-understandable primitive skills that can be flexibly combined to complete a wide range of tasks. Second, unlike prior work that models these skills~\textit{implicitly}, we~\textit{explicitly} assign a skill to each state using a lightweight neural network guided by VLM outputs, leading to more transparent and controllable behavior.

\subsubsection{Parameter synthesis.}
Hypernetworks~\cite{ha2016hypernetworks} are neural networks designed to generate the parameters of other networks, using context information as input. This approach provides an efficient way to model the dependency between the task and the optimal control policy~\cite{renhypogen}. For instance, HyperDistill~\cite{xiong2024distilling} uses a hypernetwork to learn policies for robots with different physical structures, achieving strong performance with minimal computational cost. Inspired by it, SDP establishes the dependency between skills and action predictions by parameterizing FFN layers in the diffusion policy. 

\section{Preliminaries}
The robotic manipulation learns a general policy that performs diverse tasks. Assume we have a set of robotic demonstrations $\mathcal{T} = \{\bm{\tau}_i\}_{i=1}^{|\mathcal{T}|}$, where each trajectory $\bm{\tau}_i = \{(\bm{s}_n, \overline{\bm{a}}_{n,k}, \bm{l}_i)\}_{n=1}^N$ contains the state $\bm{s}_n \in \mathbb{R}^{d_s}$, the action sequence $\overline{\bm{a}}_{n,k} \in \mathbb{R}^{7 \times k}$ of length $k$ starting at timestep $n$ and the high-level language instruction $\bm{l}_i$ specifying the task. The language-conditioned policy aims to train a policy $\pi_{\theta}(\overline{\bm{a}} | \bm{s}, \bm{l}): (\bm{s}_n, \bm{l}_i) \mapsto \overline{\bm{a}}_{n,k}$ that maps state $\bm{s}_n$ at timestep $n$ and the instruction $\bm{l}_i$ to a sequence of future actions.

Language-conditioned diffusion policy leverages the diffusion model to obtain the policy $\pi_{\theta}(\overline{\bm{a}} | \bm{s}, \bm{l})$.
To generate new samples from noise, based on historical state embeddings $\bar{\bm{s}}$ and the instructions $\bm{l}$, it trains a neural network $D_{\theta}$ to approximate the score function of the diffusion process by Denoising Score Matching~\cite{vincent2011connection}:
\begin{equation}\label{eq1}
    \mathcal{L}_{\text{SM}}(\theta;\bm{s},\bm{l}) = \mathbb{E}_{\sigma, \bar{\bm{a}}, \bm{\epsilon}} \bigg[ \frac{1}{\sigma_t} \big\| D_{\theta}(\bar{\bm{a}} + \bm{\epsilon}, \bar{\bm{s}}, \bm{l}, \sigma_t)  - \bar{\bm{a}}  \big\|_2^2 \bigg],
\end{equation}
where $\bm{\epsilon}$ is the noise and $\sigma_t$ is the density at step $t$. The diffusion model is trained by minimizing the average loss over state-action-instruction tuples from $\mathcal{T}$. Once $D_{\theta}$ is trained, the DDIM~\cite{zhang2022gddim} is adopted to sample the desired actions within $N_{d}$ denoising steps. We refer readers to MoDE~\cite{reuss2024efficient} for more details.

The Equation~(\ref{eq1}) directly maps the global instruction $\bm{l}$ to the local actions. We argue that the task specification is too abstract to provide an explicit instruction that guides the diffusion policy to generate precise short-term actions.

\section{Proposed Approach}
\subsection{Approach Overview}
This paper proposes the skill-conditioned diffusion policy to address the issue of imprecise executions from high-level instructions. The diagram is illustrated in Figure~\ref{framework}: the upper part predicts a primitive skill that describes the upcoming manipulations, and the lower part provides a single-skill policy that integrates the state information and generates skill-aligned actions. 
Notably, we use the vision-language representations to assign skills, thereby rendering the execution of tasks both interpretable and comprehensible to humans.

\subsection{Primitive Skill Assignment}
To perform a task based on a language instruction, a policy predicts short-term actions for each state. However, the coarse granularity of the high-level instruction may introduce ambiguity into fine-grained action generation. Instead, we propose to decompose tasks into fundamental, irreducible manipulation primitives, called \textit{primitive skills}. These skills provide precise, actionable guidance for generating accurate short-term controls.

\subsubsection{Compositional prompt ensemble (CPE).}
Following the intuition, we learn such primitive skills explicitly. Specifically, we abstract basic manipulations into eight reusable primitive skills, denoted as $P$,~\textit{i.e.,} ``roll'', ``yaw'', ``open the gripper'', ``move up'', ``translate'', ``close the gripper'', ``move down'', and ``rotate''. To better describe the state of the robot, we specially design a unified text template $``the \ robot \ arm \ is \ going \ to \ \{skill\}."$.
Inspired by the prompt ensemble in CLIP~\cite{radford2021learning}, we further propose the Compositional Prompt Ensemble to generate prompts for each skill, formulated as follows:
\begin{equation}
    P_{En} := ``the \ robot \ arm \ is \ going \ to \ \{skill\}." \otimes P 
\end{equation}
where $\otimes$ denotes the ensemble operation. After that, a frozen CLIP text encoder ${\rm CLIP}_{\rm text}(\cdot)$, followed by a MLP $f$, encodes the ensemble $P_{En}$ into the prompt embedding $\bm{p} = {f\rm(CLIP_{text}}(P_{En}))\in \mathbb{R}^{8\times C_{\text{img}}}$, where $C_{\text{img}}$ is the dimension of joint space for the skill assignment. Finally, one of $\{\bm{p}_i\}_{i=1}^8$ will be selected to guide the action generation.

Note that almost all tasks can be composed of those primitive skills, and the text template provides the general prompt for the robot's state. Thus, texts from CPE are reusable, and we pre-compute and store them for efficiency in inference. 

\subsubsection{Vision-language model.} Since CPE provides concrete prompts for each skill, we need to identify which skill will be performed. In particular, we utilize vision-language representations from visual observations and the high-level instruction as guidance for the assignment. Formally, let $\bm{I}_{s}, \bm{I}_{w}\in \mathbb{R}^{3\times H\times W}$ be the visual observations from the static and wrist cameras, respectively. They are encoded into visual embedding $f_{\text{img}}(\bm{I}_s)$ and $f_{\text{img}}(\bm{I}_w)$ by a shared image encoder $f_{\text{img}}(\cdot): \mathbb{R}^{3\times H\times W}\rightarrow \mathbb{R}^{N_{\text{img}}\times C_{\text{img}}}$, where $N_\text{img}$ and $C_\text{img}$ represent the number and dimensionality of vision tokens, respectively. At the same time, the high-level instruction $\bm{l}$ is processed by the tokenizer and word embedding layer from~\cite{xiao2023florence}, resulting in the text embeddings $f_{\text{t}}(\bm{l})\in \mathbb{R}^{N_{t}\times C_{\text{text}}}$. Then vision and text tokens are concatenated and sent to a transformer $\Phi$ to obtain the vision-language representations $\bm{z}_{vl} = \Phi([f_{\text{t}}(\bm{l}), f_{\text{img}}(\bm{I}_s), f_{\text{img}}(\bm{I}_w)])\in \mathbb{R}^{(N_{t}+2N_{\text{img}})\times C_{\text{img}}}$. We omit the extra projection on $f_{\text{t}}(\bm{l})$ for dimension alignment.

\subsubsection{Primitive skill selection.} Based on the embedding of skill prompts $\bm{p}$ and the vision-language representations $\bm{z}_{vl}$, we further devise the skill assignment module that employs a lightweight router network to select a skill for each state. In detail, we first average the token dimension of $\bm{z}_{vl}$ and obtian the variable $\bm{z}_{\text{avg}}\in \mathbb{R}^{C_{\text{img}}}$. Then an MLP layer maps $\bm{z}_{\text{avg}}$ into a logits to reflect the importance of each skill, followed by a $\operatorname{Softmax}$ function $\sigma(\cdot)$ and the $\operatorname{top-1}(\cdot)$ operation to narrow down all skills to the most suitable one:
\begin{equation}
    R(\bm{z_{vl}})=\operatorname{top-1}(\sigma(\operatorname{MLP}(\operatorname{Avg}(\bm{z}_{vl})))),
\end{equation}
A skill with the highest score is subsequently selected based on its importance in $R(\bm{z}_{vl})\in \mathbb{R}^8$. Finally, the skill embedding for each state is selected by $\bm{z}=\sum_{i=1}^{8}R(\bm{z}_{vl})_i\cdot \bm{p}_i$.

\subsubsection{Analysis.} The VQ~\cite{van2017neural} is utilized to~\textit{implicitly} learn discrete latent codes~\cite{garg2022lisa,liang2024skilldiffuser}. Differently, our SDP~\textit{explicitly} abstracts shared primitive skills across varying tasks, and the skill assignment is more human-understandable. More recently, GSC~\cite{mishra2023generative} explicitly parameterizes skills, pedicted by a DP. In contrary, our (finer grained) primitive skills (PS) can be assembled into their skills and express broader tasks.
Moreover, ours are learned in a unified model rather than separate ones, which is more efficient.

\subsection{Skill-conditioned Diffusion Policy Learning}
The ultimate goal is to predict skill-aligned actions. We propose to learn single-skill diffusion policies. We first inject state priors and then build a dependency between the assigned primitive skill and the conditional action generation.  

\subsubsection{Priors injection.} 
For each state, time steps, proprioception, visual observations, and high-level instruction are provided. Following work~\cite{doshi2024scaling}, a small MLP-based encoder is used to handle time steps and proprioception. Later, they are injected by a modified AdaLN~\cite{perez2017visual} that generates distinct modulation signals shared across all layers. Instead, for the visual and linguistic information, the output tokens of the VLM are first projected via a linear layer with RMSNorm~\cite{zhang2019root} and then injected by a Cross-Attention in each block, as shown in Figure~\ref{framework}. These operations integrate state priors and achieve conditional injection. Compared to the standard AdaLN that assigns unique parameters to each layer, ours reduces learnable parameters while maintaining performance.

\subsubsection{Skill-dependent FFN layer.} 
To build a dependency between the primitive skill and the action generation, we additionally introduce a LoRA-like~\cite{hu2022lora} feed-forward (FFN) layer to the original $\operatorname{FFN}_{\text{ori}}$.
The new FFN contains a SwishGLU activation and two matrices $\bm{W}_{\bm{z}}^1\in \mathbb{R}^{C\times 16}$ and $\bm{W}_{\bm{z}}^2\in \mathbb{R}^{16\times C}$, generated from the skill embed $\bm{z}$ by an MLP and mapping an input $\bm{x}$ from dimension of $C$ to $16$ and $16$ to $C$, respectively. The final FFN layer is thus formulated as:
\begin{equation}\label{FFN}
    \operatorname{FFN}(\bm{x}) = \bm{W}_{\bm{z}}^2(\operatorname{SwishGLU}(\bm{W}_{\bm{z}}^1\bm{x})) + \operatorname{FFN}_\text{ori}(\bm{x}),
\end{equation}
The LoRA-like FFN explicitly considers skill in feature extraction that saves memory and reduces the total parameters.

\subsubsection{Training Objective.}
We additionally adopt an orthogonal loss $\mathcal{L}_{\text{Orth}}(\theta)$ to reduce pairwise cosine similarity on $\bm{p}_{i,j}$ with a hyperparameter $\gamma$. The loss function thus becomes: 
\begin{equation}
    \mathcal{L}(\theta) = \mathcal{L}_{\text{SM}}(\theta) + \gamma\mathcal{L}_{\text{Orth}}(\theta),
\end{equation}
where $\mathcal{L}_{\text{Orth}} = \frac{1}{64}\sum_{i=1}^8\sum_{j=1}^8 \text{Cos}(\bm{p}_i, \bm{p}_j)$ and $\gamma=0.01$.

\subsubsection{Analysis.} Equation~(\ref{FFN}) can be viewed as a variant of the mixture of experts~\cite{jacobs1991adaptive}, where the first term is designed as a skill-dependent expert but the second one is a shared expert. Consequently, it constructs a single-skill diffusion policy that predicts skill-aligned actions. We call it a skill-conditioned diffusion policy in this paper.

\begin{table*}
\centering
\footnotesize
\begin{tabular}{ll|cccccc}
\toprule
\multirow{2}{*}{Train$\rightarrow$Test} & \multirow{2}{*}{Method} & \multicolumn{6}{c}{No. Instructions in a Row (1000 chains)} \\
\cmidrule(lr){3-8}
 &  & 1 & 2 & 3 & 4 & 5 & \textbf{Average Length} \\ \midrule
\multirow{6}{*}{ABCD$\rightarrow$D}
& DiffPolicy & 86.3\% & 72.7\% & 60.1\% & 51.2\% & 41.7\% & 3.16$\pm$0.06 \\
& RoboFlamingo & 96.4\% & 89.6\% & 82.4\% & 74.0\% & 66.0\% & 4.09$\pm$0.00 \\
& GR-1 & 94.9\% & 89.6\% & 84.4\% & 78.9\% & 73.1\% & 4.21$\pm$0.00 \\
& MDT & 98.6\% & 95.8\% & 91.6\% & 86.2\% & 80.1\% & 4.52$\pm$0.02 \\
& MoDE$^\dag$ & 95.4\% & 89.1\% & 83.8\% & 78.5\% & 73.4\% & 4.19$\pm$0.03 \\
& \textbf{SDP (Ours)} & \textbf{99.7}\% & \textbf{96.7}\% & \textbf{93.8}\% & \textbf{90.8}\% & \textbf{86.5}\% & \textbf{4.67$\pm$0.02} \\
\midrule
\multirow{7}{*}{ABC$\rightarrow$D}
& RT-1 & 53.3\% & 22.2\% & 9.4\% & 3.8\% & 1.3\% & 0.90$\pm$0.06 \\
& DiffPolicy & 63.5\% & 35.3\% & 19.4\% & 10.7\% & 6.4\% & 1.35$\pm$0.05 \\
& RoboFlamingo & 82.4\% & 61.9\% & 46.6\% & 33.1\% & 23.5\% & 2.47$\pm$0.00 \\
& OpenVLA$^\dag$ & 91.3\% & 77.8\% & 62.0\% & 52.1\% & 43.5\% & 3.27$\pm$0.00 \\
& GR-1 & 85.4\% & 71.2\% & 59.6\% & 49.7\% & 40.1\% & 3.06$\pm$0.00 \\
& UniVLA & 95.5\% & 85.8\% & 75.4\% & 66.9\% & 56.5\% & 3.80$\pm$0.07 \\
& SkillDiffuser & 94.4\% & 82.7\% & 72.1\% & 62.4\% & 55.4\% & 3.66$\pm$0.07 \\
& MoDE$^\dag$ & 95.8\% & 88.1\% & 79.3\% & 70.8\% & 62.4\% & 3.92$\pm$0.05 \\
& \textbf{SDP (Ours)} & \textbf{99.3}\% & \textbf{96.1}\% & \textbf{90.9}\% & \textbf{85.3}\% & \textbf{76.9}\% & \textbf{4.49$\pm$0.04} \\
\bottomrule
\end{tabular}
\caption{Performance on the CALVIN. Success rates for each task and average rollout length to complete 5 consecutive instructions are reported. $\pm0.00$ indicates methods without average performance and $^\dag$ means re-implementation using official codes.}
\label{calvin}
\end{table*}

\section{Experiments}
This section describes details of benchmarks and implementation. Comprehensive evaluations are conducted to study:
\begin{itemize}
    \item \textbf{Performance.} Can our SDP deliver strong performance compared to SOTA competitors across various settings?
    \item \textbf{Effectiveness}. How do the proposed design choices of our architecture impact final performance?
     \item \textbf{Interpretability.} How does SDP complete various tasks?
\end{itemize}

\subsection{Experiment Setup and Implementation Details}
\subsubsection{Simulated benchmarks.} 
We evaluate the proposed SDP on the~\textbf{CALVIN}~\cite{mees2022calvin} and~\textbf{LIBERO}~\cite{liu2023libero} benchmarks. The~\textbf{CALVIN} consists of four distinct scene configurations (splits A-D), with 34 distinct tasks of 24,000 language-annotated demonstrations. In our study, we adopt the challenging evaluation setting of ABC$\rightarrow$D, wherein policies are trained using demonstrations from environments A, B, and C, and zero-shot evaluated in environment D, and ABCD$\rightarrow$D. The evaluation protocol comprises a test set of 1,000 unique instruction chains, each consisting of five consecutive tasks. The performance is measured by success rates on sequences of 1-5 consecutive tasks and the average length of completed task sequences. The~\textbf{LIBERO}~\cite{liu2023libero} comprises multiple task suites reflecting different aspects of robotic manipulation. Our experiments focus on supervised fine-tuning within the target suite, including LIBERO-Spatial for spatial relationships, LIBERO-Object for manipulation on various objects, LIBERO-Goal for varying objectives, and LIBERO-Long for extended task duration, each consisting of 10 tasks with 50 human-teleoperated demonstrations per task.

\begin{table}
\small
  \begin{center}
  \setlength{\tabcolsep}{2.0pt}{
  \begin{tabular}{@{}@{\extracolsep{\fill}}c|cccc|c@{}}
    \toprule
    Method & Spatial & Object & Goal & Long & Average \\
    \midrule
    DiffPolicy & 78.3$\pm$0.0 & 92.5$\pm$0.0 & 68.3$\pm$0.0 & 50.5$\pm$0.0 & 72.4$\pm$0.0 \\
    Octo       & 78.9$\pm$1.0 & 85.7$\pm$0.9 & 84.6$\pm$0.9 & 51.1$\pm$1.3 & 75.1$\pm$0.6 \\
    MDT        & 78.5$\pm$0.0 & 87.5$\pm$0.0 & 73.5$\pm$0.0 & 64.8$\pm$0.0 & 76.1$\pm$0.0 \\
    OpenVLA    & 84.7$\pm$0.9 & 88.4$\pm$0.8 & 79.2$\pm$1.0 & 53.7$\pm$1.3 & 76.5$\pm$0.6 \\
    MaIL       & 74.3$\pm$0.0 & 90.1$\pm$0.0 & 81.8$\pm$0.0 & 78.6$\pm$0.0 & 83.5$\pm$0.0 \\
    UniActions & 65.0$\pm$0.0 & 78.0$\pm$0.0 & 68.0$\pm$0.0 & 47.0$\pm$0.0 & 64.5$\pm$0.0 \\
    UniVLA     & 95.2$\pm$0.0 & 95.4$\pm$0.0 & 91.9$\pm$0.0 & 87.5$\pm$0.0 & 92.5$\pm$0.0 \\
    \midrule
    \textbf{Ours} & \textbf{98.3$\pm$1.3} & \textbf{99.8$\pm$0.4} & \textbf{95.6$\pm$0.5} & \textbf{93.8$\pm$0.8} & \textbf{96.9$\pm$0.7} \\
    \bottomrule
  \end{tabular}}
  \end{center}
  \caption{Success rate (\%) on the LIBERO across four suites. Zero standard deviation indicates no average performance.}
  \label{libero}
\end{table}

\subsubsection{Real-world evaluation.} 
We design 9 tasks to evaluate the capacities of multi-task learning and visual generalization. 30 trajectories are collected for each task via a 6-DoF Lebai robot arm. The average success rate over 20 trials is reported.
\begin{itemize}
    \item \textbf{Multi-task Learning.} 1)~\textit{spatial awareness} (Pick up the lemon and put it into the pan; Open the microwave and put the chips into it). 2)~\textit{tool usage} (Sweep the cube into the dustpan; Stir water in the bowl with the spoon in the cup). 3)~\textit{semantic understanding} (Pour water from a cup into the bowl; Stack the yellow cube on another cube).
    \item \textbf{Visual Generalization.} It includes two aspects: 1) operating on unseen objects (an apple or a banana). 2) Picking and putting a lemon with~\textit{complex distractors}.
\end{itemize}

\subsubsection{Implementation Details.}
We build our SDP on a 12-block Diffusion Transformers~\cite{peebles2023scalable} and pre-trained on the OpenX~\cite{vuong2023open} following~\cite{mees2024octo}. For the simulated tasks, the model is fine-tuned on 4 A100 GPUs for 40 epochs, with AdamW as the optimizer and a learning rate of $10^{-4}$. The batch size is set to 64, and images from the static and wrist cameras are resized to $224\times224$. $N_d=4$ denoising steps are used to generate actions, and we report the average performance of overall tasks over 3 seeds. For real-world evaluation, we only use images from the static camera and train the model for 200 epochs. All results are averaged over 20 trials. Baselines are fine-tuned on real-world data with default hyperparameters.

\begin{figure*}[t]
\centering
\includegraphics[width=0.9\textwidth]{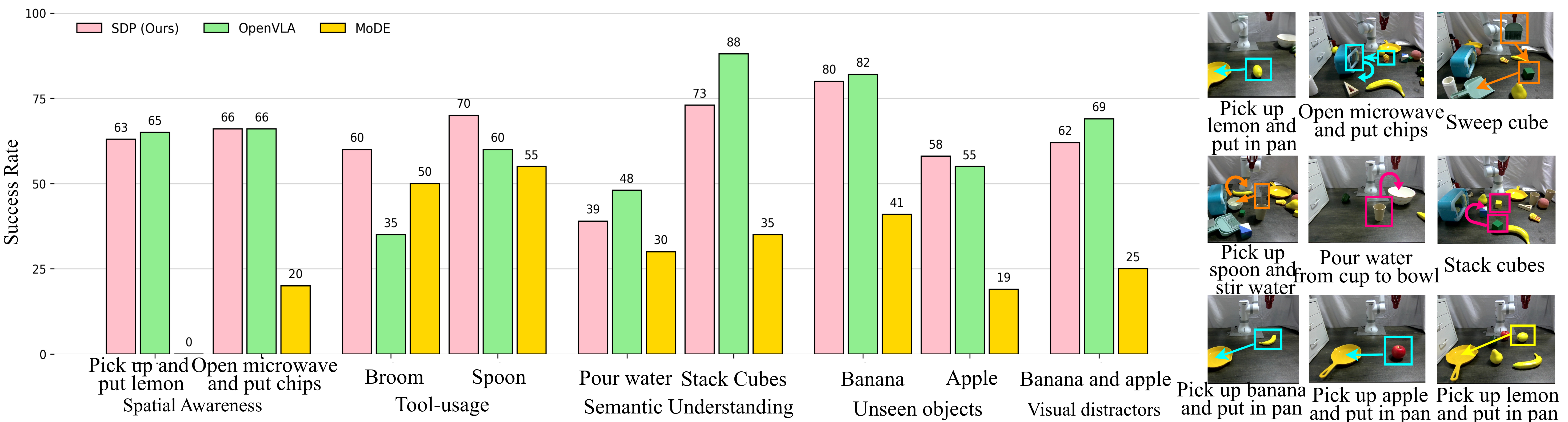}
\caption{Task success rates (\%) on real-world robot manipulation tasks. We specially designed 9 tasks (see the right figure) to evaluate two aspects of policy ability: multi-task learning (the first six tasks) and visual generalization (the last three tasks). In the visual generalization setting, we further investigate the generalization to unseen objects (an apple and a banana) and the robustness to visual distractors. The proposed SDP (pink) consistently outperforms baselines (green and orange), demonstrating better generalization across tasks and objects as well as robustness to distractors.}
\label{realtasks}
\end{figure*}

\begin{table}
\footnotesize
  \begin{center}
  \setlength{\tabcolsep}{2.0pt}{
  \begin{tabular}{@{}@{\extracolsep{\fill}}l|cccc|c@{}}
    \toprule
    \multicolumn{1}{c|}{Method} & ABCD$\rightarrow$D & ABC$\rightarrow$D & LIBERO-Long \\
    \midrule
    Baseline (DP)    & 1.98$\pm$0.09 & 1.13$\pm$0.02 & 50.5$\pm$0.5\% \\
    + Cross Atten.   & 4.09$\pm$0.07 & 3.59$\pm$0.03 & 81.0$\pm$0.8\% \\
    + Prior Injection & 4.30$\pm$0.07 & 4.01$\pm$0.04 & 86.0$\pm$0.3\% \\
    + Skill Abs.     & 4.51$\pm$0.07 & 4.32$\pm$0.07 & 91.5$\pm$0.7\% \\
    + CPE    & \textbf{4.67$\pm$0.02} & \textbf{4.49$\pm$0.05} & \textbf{93.8$\pm$0.8}\% \\
    \midrule
    \multicolumn{4}{c}{(a) Study on key components.}\\
    \midrule
    \multicolumn{1}{c}{Strategy} & ABCD$\rightarrow$D & ABC$\rightarrow$D & LIBERO-Long \\
    \midrule
    \multicolumn{1}{c}{Addition}      & 4.34$\pm$0.02 & 4.12$\pm$0.04 & 90.9$\pm$0.3\%\\  
    \multicolumn{1}{c}{Concatenation} & 4.41$\pm$0.04 & 4.24$\pm$0.06 & 91.8$\pm$0.5\% \\
    \multicolumn{1}{c}{FiLM}         & 4.49$\pm$0.03 & 4.31$\pm$0.02 & 92.5$\pm$0.6\% \\ 
    \multicolumn{1}{c}{Eq.~(\ref{FFN})} & \textbf{4.67$\pm$0.02} & \textbf{4.49$\pm$0.05} & \textbf{93.8$\pm$0.8}\% \\
    \bottomrule
    \multicolumn{4}{c}{(b) Study on strategy of skill conditioning.}\\
  \end{tabular}}
  \end{center}
  \caption{Ablations on the CALVIN and LIBERO-Long.}
  \label{ablation}
\end{table}

\subsection{Performance on Simulated Robotic Manipulation}
\subsubsection{Baselines.} 
We adopt state-of-the-art diffusion policies that report results for the CALVIN benchmark as baselines, including diffusion policy~\cite{chi2023diffusion} with CNN backbone (DiffPolicy), Octo~\cite{mees2024octo}, MDT~\cite{reuss2407multimodal}, and MoDE~\cite{reuss2024efficient}. Octo~\cite{mees2024octo} employs a unified action representation to handle heterogeneous action spaces. MDT leverages diffusion models to generate flexible action sequences conditioned on multimodal goals. MoDE combines sparse experts with a noise-conditioned self-attention mechanism to achieve more effective denoising across different noise levels. Additional baselines include current state-of-the-art VLA policies. They involve RoboFlamingo~\cite{li2023vision}, GR-1~\cite{wu2023unleashing}, OpenVLA~\cite{kim2024openvla}, and recent UniVLA~\cite{bu2025learning}. RoboFlamingo introduces alternative VLAs that use continuous action head predictions instead of discrete ones. GR-1 learns to predict future frames and actions after pre-training. OpenVLA pretrains on large-scale datasets to enable generalist robotic policies. UniVLA derives task-centric action representations from videos with a latent action model. For the LIBERO, MaIL~\cite{jia2024mail}, and UniActions~\cite{zheng2025universal} are additionally compared.

\subsubsection{Performance on the CALVIN.} 
Results in Table~\ref{calvin} demonstrate that the proposed SDP consistently outperforms all SOTA policies on both challenges. Additionally, SDP only employs four denoising steps for action generation, significantly fewer than the ten steps in diffusion-based baselines like MDT and MoDE. Specifically, on the ABCD$\rightarrow$D setting, SDP surpasses the prior state-of-the-art MDT and MoDE by a considerable margin. On the challenging ABC$\rightarrow$D setting, SDP achieves a $76.9\%$ success rate for completing all five tasks in sequence, surpassing the previous best method, MoDE by $14.5\%$, and recent UniVLA by $20.4\%$. The average number of consecutively completed tasks increases from UniVLA’s $3.80$ to $4.49$. These results not only confirm that SDP provides strong performance, but also demonstrate its ability to generalize to unseen environment settings and tackle long-horizon manipulation tasks.

\subsubsection{Performance on the LIBERO.} 
As shown in Table~\ref{libero}, our SDP demonstrates exceptional performance across all four evaluation suites, achieving high completion rates and significantly outperforming strong baselines, including MaIL and UniVLA. Notably, SDP is the only policy exceeding the success rate of $90\%$ on the LIBERO-Long suite, while other generalist approaches struggle with complex and long-horizon tasks, with only the recent UniVLA achieving competitive performance. What's more, SDP achieves an average performance of $96.9\%$, surpassing diffusion-based MDT and UniVLA by margins of $13.4\%$ and $4.4\%$, respectively. Overall, the proposed SDP demonstrates versatility and robustness across a range of robotic manipulation scenarios, leading to a new state-of-the-art on the LIBERO benchmark.

\subsection{Performance on Real-world Robot Manipulation}
\subsubsection{Baselines.} 
We compare SDP with the SOTA MoDE, employing the MoE structure, and the representative OpenVLA with a large auto-regressive architecture. 

\begin{figure*}
     \centering
    \includegraphics[width=0.98\linewidth]{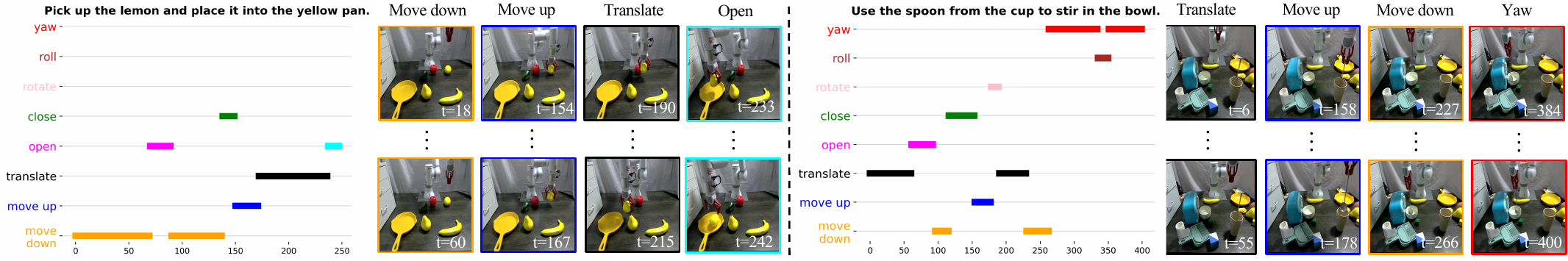} 
    \caption{Visualizations on assigned skills. The left plots draw assigned skills at each timestep, where the horizontal axis denotes the timestep, and the vertical axis corresponds to different primitive skills. Images on the right correspond to the observations by performing the skills.
    SDP learns primitive skills during training and composes them to accomplish complex tasks in inference.}\label{vis}
\end{figure*}

\subsubsection{Multi-task learning.}
The evaluation results are shown in Figure~\ref{realtasks}. The proposed SDP consistently achieves the best performance, demonstrating a clear advantage in spatial awareness, tool usage, and semantic understanding. Notably, on complex tasks, such as ``Open microwave and put chips'' and ``Pour water'', SDP outperforms other methods by a significant margin, indicating its superior ability to learn and generalize across diverse manipulation tasks. This further highlights the effectiveness of our SDP in handling complex and varied tasks within a multi-task learning scenario.

\subsubsection{Visual generalizability.}
This setting evaluates 1) the ability to manipulate objects unseen before, and 2) the robustness to irrelevant objects, known as visual distractors. 
For the former, we introduce a previously unseen apple and a banana. For the latter, we repeat the task with a lemon, but add more objects nearby to serve as distractors. The corresponding results are shown in Figure~\ref{realtasks}. The proposed SDP (pink) is capable of manipulating the apple, while baselines  (green and orange) struggle with picking and placing it. This is because SDP decomposes a task and produces skill-aware actions aligned with the task on the lemon, which is similar in shape to the apple. However, when faced with a banana, a shape not seen during training, the performance drops, indicating that generalization is more challenging for unfamiliar shapes. More importantly, the presence of visual distractors slightly impacts the success rate (from $75\%$ to $65\%$), while baselines are confused by visual distractors and perform poorly, which highlights the strong robustness of SDP.

\subsection{Effectiveness of Design Choices}

\subsubsection{Study on key components.} 
SDP has several key components: prior injection (including cross-attention and the AdaLN), skill abstraction (used by skill-dependent FFN), and compositional prompt ensemble (CPE). Table~\ref{ablation} (a) evaluates their contribution on the CALVIN and the LIBERO-Long suite. First of all, the baseline has a relatively low performance across all tasks. On the one hand, incorporating vision-language information by cross-attention significantly boosts the results, indicating its effectiveness in feature injection. Further adding other information via AdaLN continues to improve performance, demonstrating the benefit of leveraging prior knowledge. On the other hand, the introduction of skill abstraction leads to additional gains, particularly on the LIBERO-Long suite, where the score increases by $5.5\%$. Finally, based on the above structure, ensembling the compositional prompts achieves the best performance. All these results validate the importance of each component.

\begin{table}
\footnotesize
\centering
\setlength{\tabcolsep}{1.6pt}{
\begin{tabular}{ccccccc}
\toprule
& \multicolumn{1}{r}{\#Params} & FLOPS & Infer. Time & ABC$\rightarrow$D & ABCD$\rightarrow$D \\
\midrule
\multicolumn{1}{l}{Diff-P-T} & 286M & 36.3G & 22.1ms & 1.13$\pm$0.02 & 1.98$\pm$0.09 \\
\multicolumn{1}{l}{MoDE}     & 780M & 57.4G & 30.5ms & 3.92$\pm$0.07 & 4.19$\pm$0.03\\
\multicolumn{1}{l}{\textbf{Ours}} & 1017M & 74.5G & 45.1ms & 4.49$\pm$0.05 & 4.67$\pm$0.02 \\
\bottomrule
\end{tabular}}
\caption{Complexity analysis on the CALVIN benchmark.}\label{cost}
\end{table}

\subsubsection{Study on skill conditioning.}
The skill-dependent FFN helps construct the dependency between the assigned skill and the action prediction. We investigate different strategies for conditioning, including element-wise addition, channel concatenation, and FiLM~\cite{perez2018film}. As listed in Table~\ref{ablation} (b), our modeling in Equation (\ref{FFN}), parameterizing the FFN layers by the assigned skills, consistently outperforms others across all settings. Specifically, our approach achieves the highest performance on the LIBERO-Long suite, surpassing them by $2.9\%$, $2.0\%$, and $1.3\%$, respectively. Similar findings is observed in the other two. They demonstrate that the effectiveness of our strategy for complex tasks.

\subsubsection{Complexity analysis.} 
We analyze the training and deployment cost in Table~\ref{cost}. Compared to other diffusion-based policies, our SDP has a larger model size and computational cost, but consistently outperforms them by a clear margin across all tasks, with a negligible increase in inference time of 14.6ms. These comparisons demonstrate the effectiveness and efficiency of design choices despite extra overhead. 

\subsection{Visualization Analysis}
Figure~\ref{vis} visualizes the assigned skill at each timestep (left plot) and corresponding observations (right part) from conducting the skill, with the color of borders matching the skill. It is observed that SDP learns to assign reusable skills during training and sequentially composes them to accomplish the overall goals in inference. At the lower level, a diffusion model is conditioned on these skills to produce control signals that enable the desired manipulations, thereby completing complex tasks. Although the skill is assigned without explicit supervision, the visual observations are well aligned with the assigned skills, which demonstrates both the effectiveness and interpretability of our method. 
More visualizations can be found in the supplementary material.

\section{Conclusion}
This paper presents SDP, a skill-conditioned diffusion policy that integrates skill learning with conditional diffusion planning. It abstracts primitive skills from different tasks and assigns the appropriate one to guide the action generation. Experiments on both simulated and real-world tasks demonstrate the generalization and robustness, and extensive studies further validate its effectiveness and interpretability. 

\section{Acknowledgments}
The project is supported in part by the Research Grants Council (RGC) of the Hong Kong SAR through the General Research Fund (17203023), the Collaborative Research Fund (C5052-23G), and the NSFC$/$RGC Collaborative Research Scheme (CRS$\_$HKU703$/$24), and in part by UBTECH Robotics. 
The research work described in this paper was conducted while Zhihao Gu was a Postdoc of Prof. Dong Xu in the JC STEM Lab of Multimedia and Machine Learning, funded by the Hong Kong Jockey Club Charities Trust.

\bibliography{aaai2026}

@article{chi2023diffusion,
  title={Diffusion policy: Visuomotor policy learning via action diffusion},
  author={Chi, Cheng and Xu, Zhenjia and Feng, Siyuan and Cousineau, Eric and Du, Yilun and Burchfiel, Benjamin and Tedrake, Russ and Song, Shuran},
  journal={The International Journal of Robotics Research},
  volume={44},
  number={10-11},
  pages={1684--1704},
  year={2025},
  publisher={Sage Publications Sage UK: London, England}
}

@article{reuss2407multimodal,
  title={Multimodal diffusion transformer: Learning versatile behavior from multimodal goals},
  author={Reuss, Moritz and Ya{\u{g}}murlu, {\"O}mer Erdin{\c{c}} and Wenzel, Fabian and Lioutikov, Rudolf},
  journal={arXiv:2407.05996},
  year={2024}
}

@article{mandlekar2021matters,
  title={What matters in learning from offline human demonstrations for robot manipulation},
  author={Mandlekar, Ajay and Xu, Danfei and Wong, Josiah and Nasiriany, Soroush and Wang, Chen and Kulkarni, Rohun and Fei-Fei, Li and Savarese, Silvio and Zhu, Yuke and Mart{\'\i}n-Mart{\'\i}n, Roberto},
  journal={arXiv:2108.03298},
  year={2021}
}

@article{shafiullah2022behavior,
  title={Behavior transformers: Cloning $ k $ modes with one stone},
  author={Shafiullah, Nur Muhammad and Cui, Zichen and Altanzaya, Ariuntuya Arty and Pinto, Lerrel},
  journal={Advances in neural information processing systems},
  volume={35},
  pages={22955--22968},
  year={2022}
}

@inproceedings{florence2022implicit,
  title={Implicit behavioral cloning},
  author={Florence, Pete and Lynch, Corey and Zeng, Andy and Ramirez, Oscar A and Wahid, Ayzaan and Downs, Laura and Wong, Adrian and Lee, Johnny and Mordatch, Igor and Tompson, Jonathan},
  booktitle={Conference on robot learning},
  pages={158--168},
  year={2022},
}

@article{wu2020spatial,
  title={Spatial action maps for mobile manipulation},
  author={Wu, Jimmy and Sun, Xingyuan and Zeng, Andy and Song, Shuran and Lee, Johnny and Rusinkiewicz, Szymon and Funkhouser, Thomas},
  journal={arXiv:2004.09141},
  year={2020}
}

@article{ho2020denoising,
  title={Denoising diffusion probabilistic models},
  author={Ho, Jonathan and Jain, Ajay and Abbeel, Pieter},
  journal={Advances in neural information processing systems},
  volume={33},
  pages={6840--6851},
  year={2020}
}

@inproceedings{mokady2023null,
  title={Null-text inversion for editing real images using guided diffusion models},
  author={Mokady, Ron and Hertz, Amir and Aberman, Kfir and Pritch, Yael and Cohen-Or, Daniel},
  booktitle={Proceedings of the IEEE/CVF conference on computer vision and pattern recognition},
  pages={6038--6047},
  year={2023}
}

@article{reuss2024efficient,
  title={Efficient diffusion transformer policies with mixture of expert denoisers for multitask learning},
  author={Reuss, Moritz and Pari, Jyothish and Agrawal, Pulkit and Lioutikov, Rudolf},
  journal={arXiv:2412.12953},
  year={2024}
}

@article{wang2407sparse,
  title={Sparse diffusion policy: A sparse, reusable, and flexible policy for robot learning},
  author={Wang, Yixiao and Zhang, Yifei and Huo, Mingxiao and Tian, Ran and Zhang, Xiang and Xie, Yichen and Xu, Chenfeng and Ji, Pengliang and Zhan, Wei and Ding, Mingyu and others},
  journal={arXiv:2407.01531},
  year={2024}
}

@inproceedings{ha2023scaling,
  title={Scaling up and distilling down: Language-guided robot skill acquisition},
  author={Ha, Huy and Florence, Pete and Song, Shuran},
  booktitle={Conference on Robot Learning},
  pages={3766--3777},
  year={2023},
  organization={PMLR}
}

@article{mees2024octo,
  title={Octo: An open-source generalist robot policy},
  author={Team, Octo Model and Ghosh, Dibya and Walke, Homer and Pertsch, Karl and Black, Kevin and Mees, Oier and Dasari, Sudeep and Hejna, Joey and Kreiman, Tobias and Xu, Charles and others},
  journal={arXiv:2405.12213},
  year={2024}
}

@article{ze20243d,
  title={3d diffusion policy: Generalizable visuomotor policy learning via simple 3d representations},
  author={Ze, Yanjie and Zhang, Gu and Zhang, Kangning and Hu, Chenyuan and Wang, Muhan and Xu, Huazhe},
  journal={arXiv:2403.03954},
  year={2024}
}

@article{song2025survey,
  title={A survey on diffusion policy for robotic manipulation: Taxonomy, analysis, and future directions},
  author={Song, Mingchen and Deng, Xiang and Zhou, Zhiling and Wei, Jie and Guan, Weili and Nie, Liqiang},
  journal={Authorea Preprints},
  year={2025},
  publisher={Authorea}
}

@article{wang2024scaling,
  title={Scaling proprioceptive-visual learning with heterogeneous pre-trained transformers},
  author={Wang, Lirui and Chen, Xinlei and Zhao, Jialiang and He, Kaiming},
  journal={Advances in neural information processing systems},
  volume={37},
  pages={124420--124450},
  year={2024}
}

@article{ye2024latent,
  title={Latent action pretraining from videos},
  author={Ye, Seonghyeon and Jang, Joel and Jeon, Byeongguk and Joo, Sejune and Yang, Jianwei and Peng, Baolin and Mandlekar, Ajay and Tan, Reuben and Chao, Yu-Wei and Lin, Bill Yuchen and others},
  journal={arXiv:2410.11758},
  year={2024}
}

@inproceedings{liang2024skilldiffuser,
  title={Skilldiffuser: Interpretable hierarchical planning via skill abstractions in diffusion-based task execution},
  author={Liang, Zhixuan and Mu, Yao and Ma, Hengbo and Tomizuka, Masayoshi and Ding, Mingyu and Luo, Ping},
  booktitle={Proceedings of the IEEE/CVF Conference on Computer Vision and Pattern Recognition},
  pages={16467--16476},
  year={2024}
}

@article{ren2024diffusion,
  title={Diffusion policy policy optimization},
  author={Ren, Allen Z and Lidard, Justin and Ankile, Lars L and Simeonov, Anthony and Agrawal, Pulkit and Majumdar, Anirudha and Burchfiel, Benjamin and Dai, Hongkai and Simchowitz, Max},
  journal={arXiv:2409.00588},
  year={2024}
}

@article{ha2016hypernetworks,
  title={Hypernetworks},
  author={Ha, David and Dai, Andrew and Le, Quoc V},
  journal={arXiv:1609.09106},
  year={2016}
}

@inproceedings{rombach2022high,
  title={High-resolution image synthesis with latent diffusion models},
  author={Rombach, Robin and Blattmann, Andreas and Lorenz, Dominik and Esser, Patrick and Ommer, Bj{\"o}rn},
  booktitle={Proceedings of the IEEE/CVF conference on computer vision and pattern recognition},
  pages={10684--10695},
  year={2022}
}

@inproceedings{jo2022score,
  title={Score-based generative modeling of graphs via the system of stochastic differential equations},
  author={Jo, Jaehyeong and Lee, Seul and Hwang, Sung Ju},
  booktitle={International conference on machine learning},
  pages={10362--10383},
  year={2022},
  organization={PMLR}
}

@inproceedings{ni2024generate,
  title={Generate subgoal images before act: Unlocking the chain-of-thought reasoning in diffusion model for robot manipulation with multimodal prompts},
  author={Ni, Fei and Hao, Jianye and Wu, Shiguang and Kou, Longxin and Liu, Jiashun and Zheng, Yan and Wang, Bin and Zhuang, Yuzheng},
  booktitle={Proceedings of the IEEE/CVF Conference on Computer Vision and Pattern Recognition},
  pages={13991--14000},
  year={2024}
}

@article{singh2022progprompt,
  title={Progprompt: Generating situated robot task plans using large language models},
  author={Singh, Ishika and Blukis, Valts and Mousavian, Arsalan and Goyal, Ankit and Xu, Danfei and Tremblay, Jonathan and Fox, Dieter and Thomason, Jesse and Garg, Animesh},
  journal={arXiv:2209.11302},
  year={2022}
}

@article{garg2022lisa,
  title={Lisa: Learning interpretable skill abstractions from language},
  author={Garg, Divyansh and Vaidyanath, Skanda and Kim, Kuno and Song, Jiaming and Ermon, Stefano},
  journal={Advances in Neural Information Processing Systems},
  volume={35},
  pages={21711--21724},
  year={2022}
}

@article{mete2024quest,
  title={Quest: Self-supervised skill abstractions for learning continuous control},
  author={Mete, Atharva and Xue, Haotian and Wilcox, Albert and Chen, Yongxin and Garg, Animesh},
  journal={Advances in Neural Information Processing Systems},
  volume={37},
  pages={4062--4089},
  year={2024}
}

@article{xiong2024distilling,
  title={Distilling morphology-conditioned hypernetworks for efficient universal morphology control},
  author={Xiong, Zheng and Vuorio, Risto and Beck, Jacob and Zimmer, Matthieu and Shao, Kun and Whiteson, Shimon},
  journal={arXiv:2402.06570},
  year={2024}
}

@inproceedings{renhypogen,
  title={HyPoGen: Optimization-Biased Hypernetworks for Generalizable Policy Generation},
  author={Ren, Hanxiang and Sun, Li and Wang, Xulong and Zhou, Pei and Wu, Zewen and Dong, Siyan and Zou, Difan and Zheng, Youyi and Yang, Yanchao},
  booktitle={The Thirteenth International Conference on Learning Representations},
  year={2025}
}

@article{vincent2011connection,
  title={A connection between score matching and denoising autoencoders},
  author={Vincent, Pascal},
  journal={Neural computation},
  volume={23},
  number={7},
  pages={1661--1674},
  year={2011},
  publisher={MIT Press}
}

@article{zhang2022gddim,
  title={gddim: Generalized denoising diffusion implicit models},
  author={Zhang, Qinsheng and Tao, Molei and Chen, Yongxin},
  journal={arXiv:2206.05564},
  year={2022}
}

@inproceedings{xiao2023florence,
  title={Florence-2: Advancing a unified representation for a variety of vision tasks},
  author={Xiao, Bin and Wu, Haiping and Xu, Weijian and Dai, Xiyang and Hu, Houdong and Lu, Yumao and Zeng, Michael and Liu, Ce and Yuan, Lu},
  booktitle={Proceedings of the IEEE/CVF Conference on Computer Vision and Pattern Recognition},
  pages={4818--4829},
  year={2024}
}

@inproceedings{perez2017visual,
  title={Visual reasoning with a general conditioning layer, Courville},
  author={Perez, E and Strub, F and De Vries, H and Dumoulin, V},
  booktitle={In Proceedings of the AAAI Conference on Artificial Intelligence},
  year={2017}
}

@inproceedings{radford2021learning,
  title={Learning transferable visual models from natural language supervision},
  author={Radford, Alec and Kim, Jong Wook and Hallacy, Chris and Ramesh, Aditya and Goh, Gabriel and Agarwal, Sandhini and Sastry, Girish and Askell, Amanda and Mishkin, Pamela and Clark, Jack and others},
  booktitle={International conference on machine learning},
  pages={8748--8763},
  year={2021},
  organization={PMLR}
}

@article{van2017neural,
  title={Neural discrete representation learning},
  author={Van Den Oord, Aaron and Vinyals, Oriol and others},
  journal={Advances in neural information processing systems},
  volume={30},
  year={2017}
}

@inproceedings{peebles2023scalable,
  title={Scalable diffusion models with transformers},
  author={Peebles, William and Xie, Saining},
  booktitle={Proceedings of the IEEE/CVF international conference on computer vision},
  pages={4195--4205},
  year={2023}
}

@article{doshi2024scaling,
  title={Scaling cross-embodied learning: One policy for manipulation, navigation, locomotion and aviation},
  author={Doshi, Ria and Walke, Homer and Mees, Oier and Dasari, Sudeep and Levine, Sergey},
  journal={arXiv:2408.11812},
  year={2024}
}

@article{zhang2019root,
  title={Root mean square layer normalization},
  author={Zhang, Biao and Sennrich, Rico},
  journal={Advances in neural information processing systems},
  volume={32},
  year={2019}
}

@article{hu2022lora,
  title={Lora: Low-rank adaptation of large language models.},
  author={Hu, Edward J and Shen, Yelong and Wallis, Phillip and Allen-Zhu, Zeyuan and Li, Yuanzhi and Wang, Shean and Wang, Lu and Chen, Weizhu and others},
  journal={ICLR},
  volume={1},
  number={2},
  pages={3},
  year={2022}
}

@article{liu2023libero,
  title={Libero: Benchmarking knowledge transfer for lifelong robot learning},
  author={Liu, Bo and Zhu, Yifeng and Gao, Chongkai and Feng, Yihao and Liu, Qiang and Zhu, Yuke and Stone, Peter},
  journal={Advances in Neural Information Processing Systems},
  volume={36},
  pages={44776--44791},
  year={2023}
}

@article{mees2022calvin,
  title={Calvin: A benchmark for language-conditioned policy learning for long-horizon robot manipulation tasks},
  author={Mees, Oier and Hermann, Lukas and Rosete-Beas, Erick and Burgard, Wolfram},
  journal={IEEE Robotics and Automation Letters},
  volume={7},
  number={3},
  pages={7327--7334},
  year={2022},
  publisher={IEEE}
}

@inproceedings{vuong2023open,
  title={Open x-embodiment: Robotic learning datasets and rt-x models},
  author={Vuong, Quan and Levine, Sergey and Walke, Homer Rich and Pertsch, Karl and Singh, Anikait and Doshi, Ria and Xu, Charles and Luo, Jianlan and Tan, Liam and Shah, Dhruv and others},
  booktitle={Towards Generalist Robots: Learning Paradigms for Scalable Skill Acquisition@ CoRL2023},
  year={2023}
}

@article{bu2025learning,
  title={Learning to Act Anywhere with Task-centric Latent Actions},
  author={Bu, Qingwen and Yang, Yanting and Cai, Jisong and Gao, Shenyuan and Ren, Guanghui and Yao, Maoqing and Luo, Ping and Li, Hongyang},
  journal={arXiv:2502.14420},
  year={2025}
}

@article{li2023vision,
  title={Vision-language foundation models as effective robot imitators},
  author={Li, Xinghang and Liu, Minghuan and Zhang, Hanbo and Yu, Cunjun and Xu, Jie and Wu, Hongtao and Cheang, Chilam and Jing, Ya and Zhang, Weinan and Liu, Huaping and others},
  journal={arXiv:2311.01378},
  year={2023}
}

@article{wu2023unleashing,
  title={Unleashing large-scale video generative pre-training for visual robot manipulation},
  author={Wu, Hongtao and Jing, Ya and Cheang, Chilam and Chen, Guangzeng and Xu, Jiafeng and Li, Xinghang and Liu, Minghuan and Li, Hang and Kong, Tao},
  journal={arXiv:2312.13139},
  year={2023}
}

@inproceedings{jia2024mail,
  title={Mail: Improving imitation learning with selective state space models},
  author={Jia, Xiaogang and Wang, Qian and Donat, Atalay and Xing, Bowen and Li, Ge and Zhou, Hongyi and Celik, Onur and Blessing, Denis and Lioutikov, Rudolf and Neumann, Gerhard},
  booktitle={8th Annual Conference on Robot Learning},
  year={2024}
}

@article{kim2024openvla,
  title={Openvla: An open-source vision-language-action model},
  author={Kim, Moo Jin and Pertsch, Karl and Karamcheti, Siddharth and Xiao, Ted and Balakrishna, Ashwin and Nair, Suraj and Rafailov, Rafael and Foster, Ethan and Lam, Grace and Sanketi, Pannag and others},
  journal={arXiv:2406.09246},
  year={2024}
}

@article{jacobs1991adaptive,
  title={Adaptive mixtures of local experts},
  author={Jacobs, Robert A and Jordan, Michael I and Nowlan, Steven J and Hinton, Geoffrey E},
  journal={Neural computation},
  volume={3},
  number={1},
  pages={79--87},
  year={1991},
  publisher={MIT Press}
}

@article{liu2024rdt,
  title={Rdt-1b: a diffusion foundation model for bimanual manipulation},
  author={Liu, Songming and Wu, Lingxuan and Li, Bangguo and Tan, Hengkai and Chen, Huayu and Wang, Zhengyi and Xu, Ke and Su, Hang and Zhu, Jun},
  journal={arXiv:2410.07864},
  year={2024}
}

@inproceedings{perez2018film,
  title={Film: Visual reasoning with a general conditioning layer},
  author={Perez, Ethan and Strub, Florian and De Vries, Harm and Dumoulin, Vincent and Courville, Aaron},
  booktitle={Proceedings of the AAAI conference on artificial intelligence},
  volume={32},
  year={2018}
}

@inproceedings{zheng2025universal,
  title={Universal actions for enhanced embodied foundation models},
  author={Zheng, Jinliang and Li, Jianxiong and Liu, Dongxiu and Zheng, Yinan and Wang, Zhihao and Ou, Zhonghong and Liu, Yu and Liu, Jingjing and Zhang, Ya-Qin and Zhan, Xianyuan},
  booktitle={Proceedings of the Computer Vision and Pattern Recognition Conference},
  pages={22508--22519},
  year={2025}
}

@inproceedings{mishra2023generative,
  title={Generative skill chaining: Long-horizon skill planning with diffusion models},
  author={Mishra, Utkarsh Aashu and Xue, Shangjie and Chen, Yongxin and Xu, Danfei},
  booktitle={Conference on Robot Learning},
  pages={2905--2925},
  year={2023},
  organization={PMLR}
}

@article{zhang2023bootstrap,
  title={Bootstrap your own skills: Learning to solve new tasks with large language model guidance},
  author={Zhang, Jesse and Zhang, Jiahui and Pertsch, Karl and Liu, Ziyi and Ren, Xiang and Chang, Minsuk and Sun, Shao-Hua and Lim, Joseph J},
  journal={arXiv:2310.10021},
  year={2023}
}

@inproceedings{hiranaka2023primitive,
  title={Primitive skill-based robot learning from human evaluative feedback},
  author={Hiranaka, Ayano and Hwang, Minjune and Lee, Sharon and Wang, Chen and Fei-Fei, Li and Wu, Jiajun and Zhang, Ruohan},
  booktitle={2023 IEEE/RSJ International Conference on Intelligent Robots and Systems (IROS)},
  pages={7817--7824},
  year={2023},
  organization={IEEE}
}

@article{dhakan2022concurrent,
  title={Concurrent skill composition using ensemble of primitive skills},
  author={Dhakan, Paresh and Kasmarik, Kathryn and Vance, Philip and Rano, Inaki and Siddique, Nazmul},
  journal={IEEE Transactions on Cognitive and Developmental Systems},
  volume={15},
  number={4},
  pages={1879--1890},
  year={2022},
  publisher={IEEE}
}

@article{liu2025skill,
  title={Skill expansion and composition in parameter space},
  author={Liu, Tenglong and Li, Jianxiong and Zheng, Yinan and Niu, Haoyi and Lan, Yixing and Xu, Xin and Zhan, Xianyuan},
  journal={arXiv:2502.05932},
  year={2025}
}

@article{dalal2021accelerating,
  title={Accelerating robotic reinforcement learning via parameterized action primitives},
  author={Dalal, Murtaza and Pathak, Deepak and Salakhutdinov, Russ R},
  journal={Advances in Neural Information Processing Systems},
  volume={34},
  pages={21847--21859},
  year={2021}
}

\end{document}